\documentclass[11pt]{article}
\usepackage{graphicx}
\usepackage{longtable,lscape}
\usepackage{amsthm}
\usepackage{amsfonts}
\usepackage{amsmath}
\usepackage{bbm}
\usepackage{float}
\usepackage{url}

\newcommand{\captionfonts}{\footnotesize}
\makeatletter  
\long\def\@makecaption#1#2{%
  \vskip\abovecaptionskip
  \sbox\@tempboxa{{\captionfonts #1: #2}}%
  \ifdim \wd\@tempboxa >\hsize
    {\captionfonts #1: #2\par}
  \else
    \hbox to\hsize{\hfil\box\@tempboxa\hfil}%
  \fi
  \vskip\belowcaptionskip}
\makeatother 
\begin{document}
\title{Quantum Interference in Cognition: Structural Aspects of the Brain}
\author{Diederik Aerts and Sandro Sozzo \vspace{0.5 cm} \\ 
        \normalsize\itshape
        Center Leo Apostel for Interdisciplinary Studies \\
        \normalsize\itshape
        Brussels Free University \\ 
        \normalsize\itshape
         Krijgskundestraat 33, 1160 Brussels, Belgium \\
        \normalsize
        E-Mails: \url{diraerts@vub.ac.be,ssozzo@vub.ac.be} \\
               }
\date{}
\maketitle
\begin{abstract}
\noindent
We identify the presence of typically quantum effects, namely {\it superposition} and {\it interference}, in what happens when human concepts are combined, and provide a quantum model in complex Hilbert space that represents faithfully experimental data measuring the situation of combining concepts. Our model shows how `interference of concepts' explains the effects of underextension and overextension when two concepts combine to the disjunction of these two concepts. This result supports our earlier hypothesis that human thought has a superposed two-layered structure, one layer consisting of {\it classical logical thought} and a superposed layer consisting of {\it quantum conceptual thought}. Possible connections with recent findings of a {\it grid-structure} for the brain are analyzed, and influences on the mind/brain relation, and consequences on applied disciplines, such as artificial intelligence and quantum computation, are considered.
\end{abstract}


\section{Introduction\label{intro}}
In recent years it has become clear that quantum structures do not only appear within situations in the micro world, but that also situations of the macro world exhibit a quantum behavior \cite{aertsaerts1994}--\cite{aertsgaborasozzoveloz2011}. Mainly in domains such as cognitive science (decision theory, concept theory), biology (evolution theory, ecology, population dynamics) and computer science (semantic theories, information retrieval, artificial intelligence), aspects have been identified where the application of classical structures is problematic while the application of quantum structures is promising. The aspects of these domains where classical theories fail, and quantum structures are successful, reveal quite systematically four specific  and very characteristic quantum effects, namely {\it interference}, {\it contextuality}, {\it emergence} and {\it entanglement}. Sometimes it has been possible to use the full quantum apparatus of linear operators in complex Hilbert space to model these effects as they appear in these situations. However, in quite some occasions a mathematical formalism more general than standard quantum mechanics in complex Hilbert space is needed. We have introduced in \cite{aertssozzo2012} a general modeling scheme for contextual emergent entangled interfering entities. In the present article we instead focus on the identification of quantum superposition and interference in cognition to explain `how' and `why' interference models the well documented effects of {\it overextension} and {\it underextension} when concepts combine in disjunction \cite{hampton1988}. Possible connections with some recent and interesting research on the structure of the brain and technological applications to symbolic artificial intelligence and computation are also presented.

Interference effects have been studied in great detail and are very common for quantum entities, the famous `double slit situation' being an archetypical example of them \cite{young1802}--\cite{ArndtNairzVos-AndreaeKellervanderZouwZeilinger1999}. Also for concepts we have studied some effects related to the phenomenon of interference in earlier work \cite{aerts2009,aertssozzo2012}, \cite{aerts2010c}--\cite{aerts2007b}. In the present article, we concentrate on the situation where two concepts, more specifically the concepts {\it Fruits} and {\it Vegetables} are combined by using the logical connective `or' into a new concept {\it Fruits or Vegetables}. Such disjunctive combinations of concepts have been studied intensively by James Hampton \cite{hampton1988}. Hampton collected experimental data from subjects being asked to estimate the typicality of a collection of exemplars with respect to {\it Fruits} and with respect to {\it Vegetables}. Then he asked the subjects also to estimate the typicality of the same exemplars with respect to the combination {\it Fruits or Vegetables}. By using the data of these experiments we identify interference between the concepts {\it Fruits} and {\it Vegetables}, and explain how this interference accounts for the effects of underextension and overextension identified by Hampton.

In Sec. \ref{interferencesuperposition} we consider the set of data collected by Hampton, and work out a quantum description modeling these data. In Sec. \ref{graphics} we illustrate the phenomenon of interference as it appears in the considered conceptual combination, and in Sec. \ref{explanation} we present an explanation for the occurrence of this quantum effect by comparing it with the interference typical of the two-slit experiment. This modeling suggests the hypothesis in Sec. \ref{layers} that a {\it quantum conceptual layer} is present in human thought which is superposed to the usually assumed {\it classical logical thought}, the former being responsible of deviations from classically expected behavior in cognition. Finally, we present in Sec. \ref{brain} a suggestion inspired by recent research where a {\it grid}, rather than a {\it neural network}, pattern, is identified in the structure of the brain \cite{brain2012}. More specifically, we put forward the hypothesis, albeit speculative, that the interference we identity between concepts, and the complex Hilbert space that we structurally use to model this interference, might contain elements that have their isomorphic counterparts in the dynamics of the brain. Aspects of the impact of this hypothesis on the modeling and formalizing of natural and artificial knowledge, as well as the implications on artificial intelligence, robotics and quantum computation, are also inquired.

\section{Fruits interfering with Vegetables\label{interferencesuperposition}}
Let us consider the two concepts {\it Fruits} and {\it Vegetables}, and their combination {\it Fruits or Vegetables}, and work out a quantum model for the data collected by J. Hampton for this situation \cite{hampton1988,aerts2010c}. The concepts {\it Fruits} and {\it Vegetables} are two exemplars of the concept {\it Food}. And we consider a collection of exemplars of {\it Food}, more specifically those listed in Tab. 1. Then we consider the following experimental situation: Subjects are asked to respond to the following three elements: {\it Question $A$}: `Choose one of the exemplars from the list of Tab. 1 that you find a good example of {\it Fruits}'. {\it Question $B$}: `Choose one of the exemplars from the list of Tab. 1 that you find a good example of {\it Vegetables}'. {\it Question $A$ or $B$}: `Choose one of the exemplars from the list of Tab. 1 that you find a good example of {\it Fruits or Vegetables}'. Then we calculate the relative frequency $\mu(A)_k$, $\mu(B)_k$ and $\mu(A\ {\rm or}\ B)_k$, i.e the number of times that exemplar $k$ is chosen divided by the total number of choices made in response to the three questions $A$, $B$ and $A\ {\rm or}\ B$, respectively, and interpret this as an estimate for the probabilities that exemplar $k$ is chosen for questions $A$, $B$ and $A\ {\rm or}\ B$, respectively. These relative frequencies are given in Tab. 1.

For example, for {\it Question $A$}, from 10,000 subjects, 359 chose {\it Almond}, hence $\mu(A)_1=0.0359$, 425 chose {\it Acorn}, hence $\mu(A)_2=0.0425$, 372 chose {\it Peanut}, hence $\mu(A)_3=0.0372$, $\ldots$, and 127 chose {\it Black Pepper}, hence $\mu(A)_{24}=0.0127$. Analogously for {\it Question $B$}, from 10,000 subjects, 133 chose {\it Almond}, hence $\mu(B)_1=0.0133$, 108 chose {\it Acorn}, hence $\mu(B)_2=0.0108$, 220 chose {\it Peanut}, hence $\mu(B)_3=0.0220$, $\ldots$, and 294 chose {\it Black Pepper}, hence $\mu(B)_{24}=0.0294$, and for {\it Question $A\ {\rm or}\ B$}, 269 chose {\it Almond}, hence $\mu(A\ {\rm or}\ B)_1=0.0269$, 249 chose {\it Acorn}, hence $\mu(A\ {\rm or}\ B)_2=0.249$, 269 chose {\it Peanut}, hence $\mu(A\ {\rm or}\ B)_3=0.269$, $\ldots$, and 222 chose {\it Black Pepper}, hence $\mu(A\ {\rm or}\ B)_{24}=0.222$.

Let us now explicitly construct a quantum mechanical model in complex Hilbert space for the pair of concepts {\it Fruit} and {\it Vegetable} and their disjunction `{\it Fruit or Vegetable}', and show that quantum interference models the experimental data gathered in \cite{hampton1988}. We represent the measurement of `a good example of' by means of a self-adjoint operator with spectral decomposition $\{M_k\ \vert\ k=1,\ldots,24\}$ where each $M_k$ is an orthogonal projection of the Hilbert space ${\cal H}$ corresponding to item $k$ from the list of items in Tab. 1. 
\begin{table}[H]
\begin{center}
\begin{tabular}{|llllllll|}
\hline 
\multicolumn{2}{|l}{} & \multicolumn{1}{l}{$\mu(A)_k$} & \multicolumn{1}{l}{$\mu(B)_k$} & \multicolumn{1}{l}{$\mu(A\ {\rm or}\ B)_k$} & \multicolumn{1}{l}{
${\mu(A)_k+\mu(B)_k \over 2}$} & \multicolumn{1}{l}{$\lambda_k$} & \multicolumn{1}{l|}{$\phi_k$} \\
\hline
\multicolumn{8}{|l|}{\it $A$=Fruits, $B$=Vegetables} \\
\hline
1 & {\it Almond} & 0.0359 & 0.0133 & 0.0269 & 0.0246 & 0.0218 & 83.8854$^\circ$ \\
2 & {\it Acorn} & 0.0425 & 0.0108 & 0.0249 & 0.0266 & -0.0214 & -94.5520$^\circ$ \\
3 & {\it Peanut} & 0.0372 & 0.0220 & 0.0269 & 0.0296 & -0.0285 & -95.3620$^\circ$ \\
4 & {\it Olive} & 0.0586 & 0.0269 & 0.0415 & 0.0428 & 0.0397 & 91.8715$^\circ$ \\
5 & {\it Coconut} & 0.0755 & 0.0125 & 0.0604 & 0.0440 & 0.0261 & 57.9533$^\circ$ \\
6 & {\it Raisin} & 0.1026 & 0.0170 & 0.0555 & 0.0598 & 0.0415 & 95.8648$^\circ$ \\
7 & {\it Elderberry} & 0.1138 & 0.0170 & 0.0480 & 0.0654 & -0.0404 & -113.2431$^\circ$ \\ 
8 & {\it Apple} & 0.1184 & 0.0155 & 0.0688 & 0.0670 & 0.0428 & 87.6039$^\circ$ \\ 
9 & {\it Mustard} & 0.0149 & 0.0250 & 0.0146 & 0.0199 & -0.0186 & -105.9806$^\circ$ \\
10 & {\it Wheat} & 0.0136 & 0.0255 & 0.0165 & 0.0195 & 0.0183 & 99.3810$^\circ$ \\ 
11 & {\it Root Ginger} & 0.0157 & 0.0323 & 0.0385 & 0.0240 & 0.0173 & 50.0889$^\circ$ \\
12 & {\it Chili Pepper} & 0.0167 & 0.0446 & 0.0323 & 0.0306 & -0.0272 &  -86.4374$^\circ$ \\ 
13 & {\it Garlic} & 0.0100 & 0.0301 & 0.0293 & 0.0200 & -0.0147 & -57.6399$^\circ$ \\
14 & {\it Mushroom} & 0.0140 & 0.0545 & 0.0604 & 0.0342 & 0.0088 & 18.6744$^\circ$ \\
15 & {\it Watercress} & 0.0112 & 0.0658 & 0.0482 & 0.0385 & -0.0254 &  -69.0705$^\circ$ \\
16 & {\it Lentils} & 0.0095 & 0.0713 & 0.0338 & 0.0404 & 0.0252 & 104.7126$^\circ$ \\
17 & {\it Green Pepper} & 0.0324 & 0.0788 & 0.0506 & 0.0556 & -0.0503 & -95.6518$^\circ$ \\
18 & {\it Yam} & 0.0533 & 0.0724 & 0.0541 & 0.0628 & 0.0615 & 98.0833$^\circ$ \\
19 & {\it Tomato} & 0.0881 & 0.0679 & 0.0688 & 0.0780 & 0.0768 & 100.7557$^\circ$ \\
20 & {\it Pumpkin} & 0.0797 & 0.0713 & 0.0579 & 0.0755 & -0.0733 & -103.4804$^\circ$  \\
21 & {\it Broccoli} & 0.0143 & 0.1284 & 0.0642 & 0.0713 & -0.0422 & -99.6048$^\circ$ \\
22 & {\it Rice} & 0.0140 & 0.0412 & 0.0248 & 0.0276 & -0.0238 & -96.6635$^\circ$ \\ 
23 & {\it Parsley} & 0.0155 & 0.0266 & 0.0308 & 0.0210 & -0.0178 & -61.1698$^\circ$ \\
24 & {\it Black Pepper} & 0.0127 & 0.0294 & 0.0222 & 0.0211 & 0.0193 & 86.6308$^\circ$ \\
\hline
\end{tabular}
\end{center}
\caption{Interference data for concepts {\it A=Fruits} and {\it B=Vegetables}. The probability of a person choosing one of the exemplars as an example of {\it Fruits} (and as an example of {\it Vegetables}, respectively), is given by $\mu(A)$ (and $\mu(B)$, respectively) for each of the exemplars. The probability of a person choosing one of the exemplars as an example of {\it Fruits or Vegetables} is $\mu(A\ {\rm or}\ B)$ for each of the exemplars. The classical probability would be given by ${\mu(A)+\mu(B) \over 2}$, and $\phi_k$ is the quantum phase angle provoking the quantum interference effect.
}
\end{table}

\noindent
The concepts {\it Fruits}, {\it Vegetables} and `{\it Fruits or Vegetables}' are represented by unit vectors $|A\rangle$, $|B\rangle$ and ${1 \over \sqrt{2}}(|A\rangle+|B\rangle)$ of the Hilbert space ${\cal H}$, where $|A\rangle$ and $|B\rangle$ are orthgonal, and ${1 \over \sqrt{2}}(|A\rangle+|B\rangle)$ is their normalized superposition. 
Following standard quantum rules we have $\mu(A)_k=\langle A|M_k|A\rangle$, $\mu(B)_k=\langle B|M_k|B\rangle$, hence
\begin{equation} 
\mu(A\ {\rm or}\ B)_k={1 \over 2}\langle A+B|M_k|A+B\rangle={1 \over 2}(\mu(A)_k+\mu(B)_k)+\Re\langle A|M_k|B\rangle, \label{muAorB}
\end{equation}
where $\Re\langle A|M_k|B\rangle$ is the interference term. Let us introduce $|e_k\rangle$ the unit vector on $M_k|A\rangle$ and $|f_k\rangle$ the unit vector on $M_k|B\rangle$, and put $\langle e_k|f_k\rangle=c_ke^{i\gamma_k}$. Then we have $|A\rangle=\sum_{k=1}^{24}a_ke^{i\alpha_k}|e_k\rangle$ and $|B\rangle=\sum_{k=1}^{24}b_ke^{i\beta_k}|f_k\rangle$, which gives
\begin{equation} \label{ABequation}
\langle A|B\rangle=(\sum_{k=1}^{24}a_ke^{-i\alpha_k}\langle e_k|)(\sum_{l=1}^{24}b_le^{i\beta_l}|f_l\rangle)=\sum_{k=1}^{24}a_kb_kc_ke^{i\phi_k}
\end{equation}
where we put $\phi_k=\beta_k-\alpha_k+\gamma_k$. Further we have $\mu(A)_k=a_k^2$, $\mu(B)_k=b_k^2$, $\langle A|M_k|B\rangle=a_kb_kc_ke^{i\phi_k}$,
which gives, by using (\ref{muAorB}),
\begin{equation} \label{muAorBequation}
\mu(A\ {\rm or}\ B)_k={1 \over 2}(\mu(A)_k+\mu(B)_k)+c_k\sqrt{\mu(A)_k\mu(B)_k}\cos\phi_k
\end{equation}
We choose $\phi_k$ such that
\begin{equation} \label{cosequation}
\cos\phi_k={2\mu(A\ {\rm or}\ B)_k-\mu(A)_k-\mu(B)_k \over 2c_k\sqrt{\mu(A)_k\mu(B)_k}}
\end{equation}
and hence (\ref{muAorBequation}) is satisfied. We now have to determine $c_k$ in such a way that $\langle A|B\rangle=0$. Recall that from $\sum_{k=1}^{24}\mu(A\ {\rm or}\ B)_k=1$ and (\ref{muAorBequation}), and with the choice of $\cos\phi_k$ that we made in (\ref{cosequation}), it follows that $\sum_{k=1}^{24}c_k\sqrt{\mu(A)_k\mu(B)_k}\cos\phi_k=0$. Taking into account (\ref{ABequation}), which gives $\langle A|B\rangle=\sum_{k=1}^{24}a_kb_kc_k(\cos\phi_k+i\sin\phi_k)$, and making use of $\sin\phi_k=\pm\sqrt{1-\cos^2\phi_k}$, we have $\langle A|B\rangle=0$ $\Leftrightarrow$ $\sum_{k=1}^{24}c_k\sqrt{\mu(A)_k\mu(B)_k}(\cos\phi_k+i\sin\phi_k)=0$ $\Leftrightarrow$ $\sum_{k=1}^{24}c_k\sqrt{\mu(A)_k\mu(B)_k}\sin\phi_k=0$ $\Leftrightarrow$
\begin{equation}
\label{conditionequation}
\sum_{k=1}^{24}\pm\sqrt{c_k^2\mu(A)_k\mu(B)_k-(\mu(A\ {\rm or}\ B)_k-{\mu(A)_k+\mu(B)_k \over 2})^2}=0 
\end{equation}
We introduce the following quantities
\begin{equation} \label{lambdak}
\lambda_k=\pm\sqrt{\mu(A)_k\mu(B)_k-(\mu(A\ {\rm or}\ B)_k-{\mu(A)_k+\mu(B)_k \over 2})^2}
\end{equation}
and choose $m$ the index for which $|\lambda_m|$ is the biggest of the $|\lambda_k|$'s. Then we take $c_k=1$ for $k\not=m$. We explain now the algorithm that we use to choose a plus or minus sign for $\lambda_k$ as defined in (\ref{lambdak}), with the aim of being able to determine $c_m$ such that (\ref{conditionequation}) is satisfied. We start by choosing a plus sign for $\lambda_m$. Then we choose a minus sign in (\ref{lambdak}) for the $\lambda_k$ for which $|\lambda_k|$ is the second biggest; let us call the index of this term $m_2$. This means that $0\le\lambda_m+\lambda_{m_2}$. For the $\lambda_k$ for which $|\lambda_k|$ is the third biggest -- let us call the index of this term $m_3$ -- we choose a minus sign in case $0\le\lambda_m+\lambda_{m_2}+\lambda_{m_3}$, and otherwise we choose a plus sign, and in this case we have $0\le\lambda_m+\lambda_{m_2}+\lambda_{m_3}$. We continue this way of choosing, always considering the next biggest $|\lambda_k|$, and hence arrive at a global choice of signs for all of the $\lambda_k$, such that $0\le\lambda_m+\sum_{k\not=m}\lambda_k$. Then we determine $c_m$ such that (\ref{conditionequation}) is satisfied, or more specifically such that
\begin{equation} \label{cmequation}
c_m=\sqrt{{(-\sum_{k\not=m}\lambda_k)^2+(\mu(A\ {\rm or}\ B)_m-{\mu(A)_m+\mu(B)_m \over 2})^2 \over \mu(A)_m\mu(B)_m}}
\end{equation}
We choose the sign for $\phi_k$ as defined in (\ref{cosequation}) equal to the sign of $\lambda_k$. The result of the specific solution that we have constructed is that we can take $M_k({\cal H})$ to be rays of dimension 1 for $k\not=m$, and $M_m({\cal H})$ to be a plane. This means that we can make our solution still more explicit. Indeed, we take ${\cal H}={\mathbb{C}}^{25}$ the canonical 25 dimensional complex Hilbert space, and make the following choices
\begin{equation} \label{vectorA}
|A\rangle=(\sqrt{\mu(A)_1},\ldots,\sqrt{\mu(A)_m},\ldots,
\sqrt{\mu(A)_{24}},0)
\end{equation}
\begin{equation}
|B\rangle=(e^{i\beta_1}\sqrt{\mu(B)_1},\ldots,c_me^{i\beta_m}\sqrt{\mu(B)_m},\ldots,
e^{i\beta_{24}}\sqrt{\mu(B)_{24}},\sqrt{\mu(B)_m(1-c_m^2)}) \label{vectorB}
\end{equation}
\begin{equation}\label{anglebetan}
\beta_m=\arccos({2\mu(A\ {\rm or}\ B)_m-\mu(A)_m-\mu(B)_m \over 2c_m\sqrt{\mu(A)_m\mu(B)_m}}) \\
\end{equation}
\begin{equation}\label{anglebetak}
\beta_k=\pm\arccos({2\mu(A\ {\rm or}\ B)_k-\mu(A)_k-\mu(B)_k \over 2\sqrt{\mu(A)_k\mu(B)_k}})
\end{equation}
where the plus or minus sign in (\ref{anglebetak}) is chosen following the algorithm we introduced for choosing the plus and minus sign for $\lambda_k$ in (\ref{lambdak}). Let us construct this quantum model for the data given in Tab. 1.
 The exemplar which gives rise to the biggest value of $|\lambda_k|$ is {\it Tomato}, and hence we choose a plus sign and get $\lambda_{19}=0.0768$. The exemplar giving rise to the second biggest value of $\lambda_k$ is {\it Pumpkin}, and hence we choose a minus sign, and get $\lambda_{20}=-0.0733$. Next comes {\it Yam}, and since $\lambda_{19}+\lambda_{20}-0.0615<0$, we choose a plus sign for $\lambda_{18}$. Next is {\it Green Pepper}, and we look at $0\le\lambda_{19}+\lambda_{20}+\lambda_{18}-0.0503$, which means that we can choose a minus sign for $\lambda_{17}$. The fifth exemplar in the row is {\it Apple}. We have $\lambda_{19}+\lambda_{20}+\lambda_{18}+\lambda_{17}-0.0428<0$, which means that we need to choose a plus sign for $\lambda_8$. Next comes {\it Broccoli} and verifying shows that we can choose a minus sign for $\lambda_{21}$. We determine in an analogous way the signs for the exemplars {\it Raisin}, plus sign, {\it Elderberry}, minus sign, {\it Olive}, plus sign, {\it Peanut}, minus sign, {\it Chili Pepper}, minus sign, {\it Coconut}, plus sign, {\it Watercress}, minus sign, {\it Lentils}, plus sign, {\it Rice}, minus sign, {\it Almond}, plus sign, {\it Acorn}, minus sign, {\it Black Pepper}, plus sign, {\it Mustard}, minus sign, {\it Wheat}, plus sign, {\it Parsley}, minus sign, {\it Root Ginger}, plus sign, {\it Garlic}, minus sign, and finally {\it Mushroom}, plus sign. In Tab. 1 we give the values of $\lambda_k$ calculated following this algorithm, and from (\ref{cmequation}) it follows that $c_{19}=0.7997$.

Making use of (\ref{vectorA}), (\ref{vectorB}), (\ref{anglebetan}) and (\ref{anglebetak}), and the values of the angles given in Tab. 1, we put forward the following explicit representation of the vectors $|A\rangle$ and $|B\rangle$ in ${\mathbb{C}}^{25}$ representing concepts {\it Fruits} and {\it Vegetables}
\begin{eqnarray}
|A\rangle&=&(0.1895, 0.2061, 0.1929, 0.2421, 0.2748, 0.3204, 0.3373, 0.3441, 0.1222, 0.1165, 0.1252, 0.1291,  \nonumber \\ 
          && 0.1002, 0.1182, 0.1059, 0.0974, 0.1800, 0.2308, 0.2967, 0.2823, 0.1194, 0.1181, 0.1245, 0.1128, 0) \nonumber \\ 
|B\rangle&=&(0.1154e^{i83.8854^\circ}, 0.1040e^{-i94.5520^\circ}, 0.1484e^{-i95.3620^\circ}, 0.1640e^{i91.8715^\circ}, 0.1120e^{i57.9533^\circ}, \nonumber \\
          && 0.1302e^{i95.8648^\circ}, 0.1302e^{-i113.2431^\circ}, 0.1246e^{i87.6039^\circ}, 0.1580e^{-i105.9806^\circ},0.1596e^{i99.3810^\circ}, \nonumber \\ 
          &&  0.1798e^{i50.0889^\circ}, 0.2112e^{-i86.4374^\circ}, 0.1734e^{-i57.6399^\circ}, 0.2334e^{i18.6744^\circ}, 0.2565e^{-i69.0705^\circ}, \nonumber \\
          && 0.2670e^{i104.7126^\circ}, 0.2806e^{-i95.6518^\circ}, 0.2690e^{i98.0833^\circ}, 0.2606e^{i100.7557^\circ}, 0.2670e^{-i103.4804^\circ}, \nonumber \\
          && 0.3584e^{-i99.6048^\circ}, 0.2031e^{-i96.6635^\circ}, 0.1630e^{-i61.1698^\circ}, 0.1716e^{i86.6308^\circ}, 0.1565). \label{interferenceangles}
\end{eqnarray}
This proves that we can model the data of \cite{hampton1988} by means of a quantum mechanical model, and such that the values of $\mu(A\ {\rm or}\ B)_k$ are determined from the values of $\mu(A)_k$ and $\mu(B)_k$ as a consequence of quantum interference effects. For each $k$ the value of $\phi_k$ in Tab. 1 gives the quantum interference phase of the exemplar number $k$.

\section{Graphics of the interference patterns\label{graphics}}
In \cite{aerts2010c} we worked out a way to `chart' the quantum interference patterns of the two concepts when combined into conjunction or disjunction. Since it helps our further analysis in the present article, we put forward this `chart' for the case of the concepts {\it Fruits} and {\it Vegetables} and their disjunction `{\it Fruits or Vegetables}'. More specifically, we represent the concepts {\it Fruits}, {\it Vegetables} and `{\it Fruits or Vegetables}' by complex valued wave functions of two real variables $\psi_A(x,y), \psi_B(x,y)$ and $\psi_{A{\rm or}B}(x,y)$. We choose $\psi_A(x,y)$ and $\psi_B(x,y)$ such that the real part for both wave functions is a Gaussian in two dimensions, which is always possible since we have to fit in only 24 values, namely the values of $\psi_A$ and $\psi_B$ for each of the exemplars of Tab. 1. The squares of these Gaussians are graphically represented in Figs. 1 and 2, and the different exemplars of Tab. 1 are located in spots such that the Gaussian distributions $|\psi_A(x,y)|^2$ and $|\psi_B(x,y)|^2$ properly model the probabilities $\mu(A)_k$ and $\mu(B)_k$ in Tab. 1 for each one of the exemplars. For example, for {\it Fruits} represented in Fig. 1, {\it Apple} is located in the center of the Gaussian, since {\it Apple} was most frequently chosen by the test subjects when asked {\it Question A}. {\it Elderberry} was the second most frequently chosen, and hence closest to the top of the Gaussian in Fig. 1.
\begin{figure}[H]
\centerline {\includegraphics[scale=0.58]{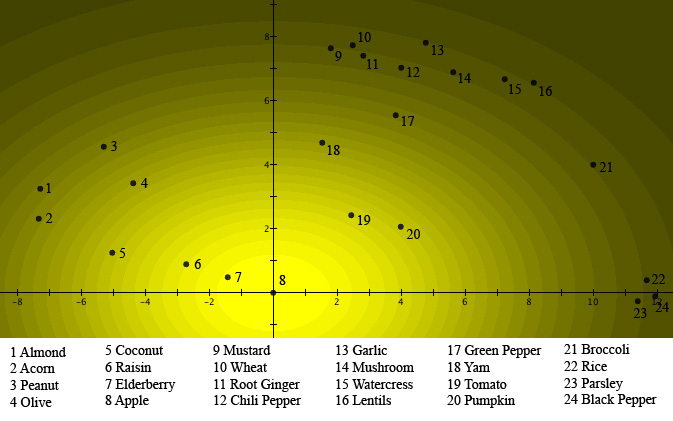}}
\caption{The probabilities $\mu(A)_k$ of a person choosing the exemplar $k$ as a `good example' of {\it Fruits} are fitted into a two-dimensional quantum wave function $\psi_A(x,y)$. The numbers are placed at the locations of the different exemplars with respect to the Gaussian probability distribution $|\psi_A(x,y)|^2$. This can be seen as a light source shining through a hole centered on the origin, and regions where the different exemplars are located. The brightness of the light source in a specific region corresponds to the probability that this exemplar will be chosen as a `good example' of {\it Fruits}.
}
\end{figure}
\begin{figure}[H]
\centerline {\includegraphics[scale=0.58]{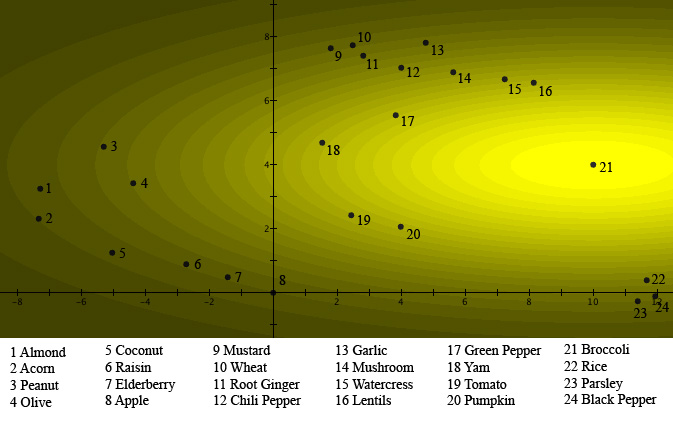}}
\caption{The probabilities $\mu(B)_k$ of a person choosing the exemplar $k$ as an example of {\it Vegetables} are fitted into a two-dimensional quantum wave function $\psi_B(x,y)$. The numbers are placed at the locations of the different exemplars with respect to the probability distribution $|\psi_B(x,y)|^2$. As in Fig. 1, it can be seen as a light source shining through a hole centered on point 21, where {\it Broccoli} is located. The brightness of the light source in a specific region corresponds to the probability that this exemplar will be chosen as a `good example' of {\it Vegetables}.
}
\end{figure}
\noindent
Then come {\it Raisin}, {\it Tomato} and {\it Pumpkin}, and so on, with {\it Garlic} and {\it Lentils} as the least chosen `good examples' of {\it Fruits}. For {\it Vegetables}, represented in Fig. 2, {\it Broccoli} is located in the center of the Gaussian, since {\it Broccoli} was the exemplar most frequently chosen by the test subjects when asked {\it Question B}. {\it Green Pepper} was the second most frequently chosen, and hence closest to the top of the Gaussian in Fig. 2. Then come {\it Yam}, {\it Lentils} and {\it Pumpkin}, and so on, with {\it Coconut} and {\it Acorn} as the least chosen `good examples' of {\it Vegetables}. Metaphorically, we could regard the graphical representations of Figs. 1 and 2 as the projections of two light sources each shining through one of two holes in a plate and spreading out their light intensity following a Gaussian distribution when projected on a screen behind the holes.
\begin{figure}[H]
\centerline {\includegraphics[scale=0.58]{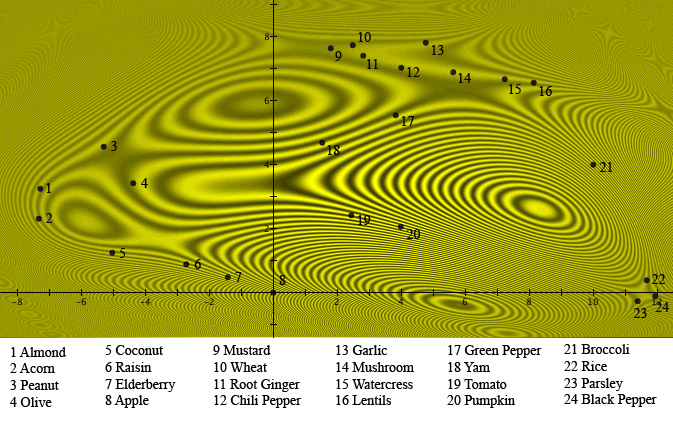}}
\caption{The probabilities $\mu(A\ {\rm or}\ B)_k$ of a person choosing the exemplar $k$ as an example of `{\it Fruits or Vegetables}' are fitted into the two-dimensional quantum wave function ${1 \over \sqrt{2}}(\psi_A(x,y)+\psi_B(x,y))$, which is the normalized superposition of the wave functions in Figs. 1 and 2. The numbers are placed at the locations of the different exemplars with respect to the probability distribution ${1 \over 2}|\psi_A(x,y)+\psi_B(x,y)|^2={1 \over 2}(|\psi_A(x,y)|^2+|\psi_B(x,y)|^2)+|\psi_A(x,y)\psi_B(x,y)|\cos\phi(x,y)$, where $\phi(x,y)$ is the quantum phase difference at $(x,y)$. The values of $\phi(x,y)$ are given in Tab. 1 for the locations of the different exemplars. The interference pattern is clearly visible.
}
\end{figure}

\noindent
The center of the first hole, corresponding to the {\it Fruits} light source, is located where exemplar {\it Apple} is at point $(0, 0)$, indicated by 8 in both figures. The center of the second hole, corresponding to the {\it Vegetables} light source, is located where exemplar {\it Broccoli} is at point (10,4), indicated by 21 in both figures.

In Fig. 3 the data for `{\it Fruits or Vegetables}' are graphically represented. This is not `just' a normalized sum of the two Gaussians of Figs. 1 and 2, since it is the probability distribution corresponding to ${1 \over \sqrt{2}}(\psi_A(x,y)+\psi_B(x,y))$, which is the normalized superposition of the wave functions in Figs. 1 and 2. The numbers are placed at the locations of the different exemplars with respect to the probability distribution ${1 \over 2}|\psi_A(x,y)+\psi_B(x,y)|^2={1 \over 2}(|\psi_A(x,y)|^2+|\psi_B(x,y)|^2)+|\psi_A(x,y)\psi_B(x,y)|\cos\phi(x,y)$, where $|\psi_A(x,y)\psi_B(x,y)|\cos\phi(x,y)$ is the interference term and $\phi(x,y)$ the quantum phase difference at $(x,y)$. The values of $\phi(x,y)$ are given in Tab. 1 for the locations of the different exemplars. The interference pattern shown in Fig. 3 is very similar to well-known interference patterns of light passing through an elastic material under stress. In our case, it is the interference pattern corresponding to `{\it Fruits or Vegetables}'. Bearing in mind the analogy with the light sources for Figs. 1 and 2, in Fig. 3 we can see the interference pattern produced when both holes are open. 

 \begin{figure}[H]
\centerline {\includegraphics[scale=0.4]{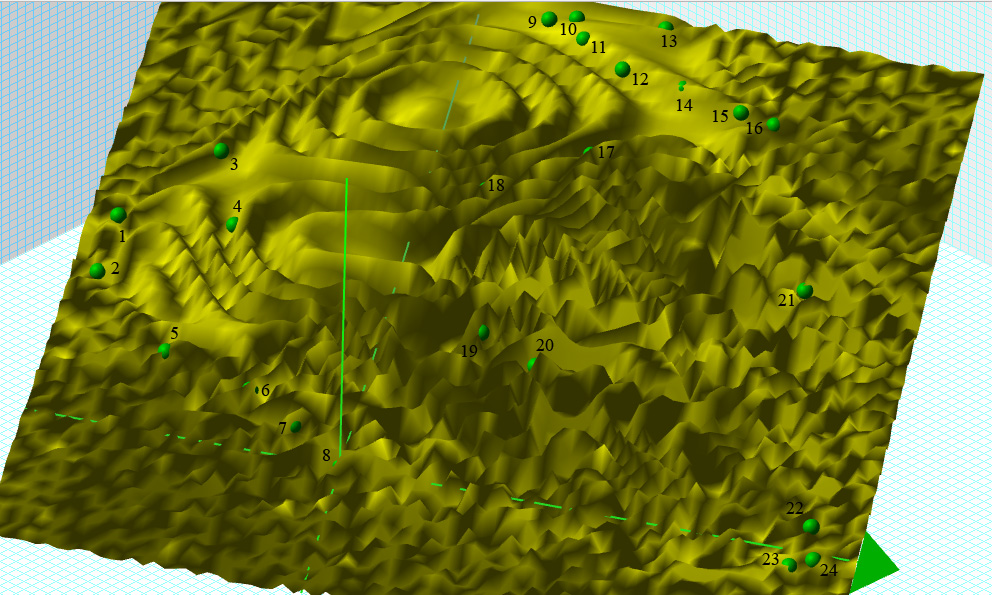}}
\caption{A three-dimensional representation of the interference landscape of the concept `{\it Fruits {\rm or} Vegetables}' as shown in Fig. 3. Exemplars are represented by little green balls, and the numbers refer to the numbering of the exemplars in Tab. 1 and in Figs. 1, 2 and 3.
}
\end{figure}
\begin{figure}[H]
\centerline {\includegraphics[scale=0.58]{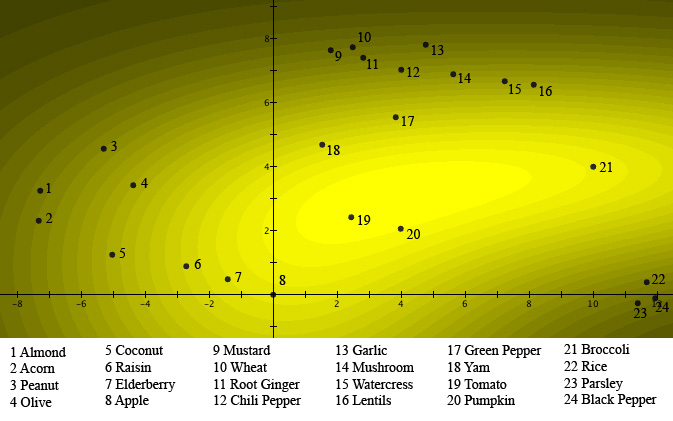}}
\caption{Probabilities $1/2(\mu(A)_k+\mu(B)_k)$, which are the probability averages for {\it Fruits} and {\it Vegetables} shown in Figs. 1 and 2. This would be the resulting pattern in case $\phi(x,y)=90^\circ$ for all exemplars. It is called the classical pattern for the situation since it is the pattern that, without interference, results from a situation where classical particles are sent through two slits. These classical values for all exemplars are given in Tab. 1.
}
\end{figure}
\noindent
Fig. 4 represents a three-dimensional graphic of the interference pattern of Fig. 3, and, for the sake of comparison, in Fig. 5, we have graphically represented the averages of the probabilities of Figs. 1 and 2, i.e. the values measured if there were no interference. For the mathematical details -- the exact form of the wave functions and the explicit calculation of the interference pattern -- and for other examples of conceptual interference, we refer to \cite{aerts2010c}.

\section{Explaining quantum interference\label{explanation}}
The foregoing section showed how the typicality data of two concepts and their disjunction are quantum mechanically modeled such that the quantum effect of interference accounts for the measured values. We also showed that it is possible to metaphorically picture the situation such that each of the concepts is represented by light passing through a hole and the disjunction of both concepts corresponds to the situation of the light passing through both holes (see Fig. 6). 
\begin{figure}[H]
\centerline {\includegraphics[scale=0.29]{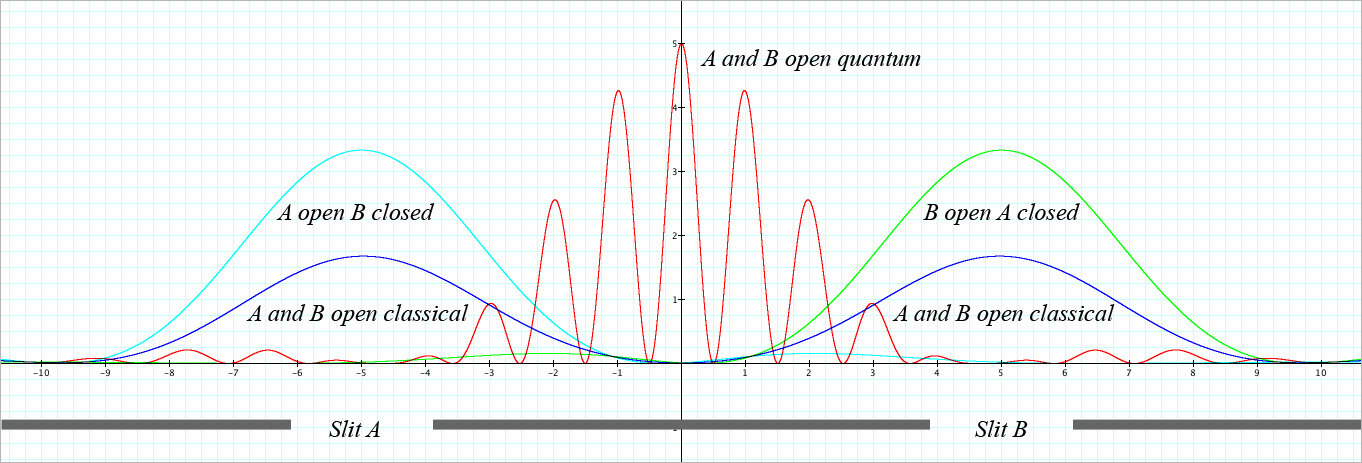}}
\caption{A typical interference pattern of a quantum two-slit situation with slits $A$ and $B$. The `{\it A open B closed}' curve represents the probability of detection of the quantum entity in case only {\it Slit A} is open; the `{\it B open A closed}' curve reflects the situation where only {\it Slit B} is open; and the `{\it A and B open classical}' curve is the average of both. The `{\it A and B open quantum}' curve represents the probability of detection of the quantum entity if both slits are open.
}
\end{figure}
\noindent
This is indeed where interference is best known from in the traditional double-slit situation in optics and quantum physics. If we apply this to our specific example by analogy, we can imagine the cognitive experiment where a subject chooses the most appropriate answer for one of the concepts, e.g., {\it Fruits}, as follows: `The photon passes with the {\it Fruits} hole open and hits a screen behind the hole in the region where the choice of the person is located'. We can do the same for the cognitive experiment where the subject chooses the most appropriate answer for the concept {\it Vegetables}. This time the photon passes with the {\it Vegetables} hole open and hits the screen in the region where the choice of the person is located. The third situation, corresponding to the choice of the most appropriate answer for the disjunction concept `{\it Fruits or Vegetables}', consists in the photon passing with both the {\it Fruits} hole and the {\it Vegetables} hole open and hitting the screen where the choice of the person is located. This third situation is the situation of interference, viz. the interference between {\it Fruits} and {\it Vegetables}. These three situations are clearly illustrated in Figs. 1, 2 and 3.

In \cite{aerts2009,aerts2007a,aerts2007b} we analyzed the origin of the interference effects that are produced when concepts are combined, and we provided an explanation that we investigated further in \cite{aertsdhooghe2009}. 

Let us now take a closer look at the experimental data and how they are produced by interference. The exemplars for which the interference is a weakening effect, i.e. where $\mu(A\ {\rm or}\ B) < 1/2(\mu(A)+\mu(B))$ or $90^\circ \le \phi$ or $\phi \le -90^\circ$, are the following: {\it Elderberry}, {\it Mustard}, {\it Lentils}, {\it Pumpkin}, {\it Tomato}, {\it Broccoli}, {\it Wheat}, {\it Yam}, {\it Rice}, {\it Raisin}, {\it Green Pepper}, {\it Peanut}, {\it Acorn} and {\it Olive}. The exemplars for which interference is a strengthening effect, i.e. where $1/2(\mu(A)+\mu(B)) < \mu(A\ {\rm or}\ B)$ or $\phi < 90^\circ$ or $-90^\circ \le \phi$, are the following: {\it Mushroom}, {\it Root Ginger}, {\it Garlic}, {\it Coconut}, {\it Parsley}, {\it Almond}, {\it Chili Pepper}, {\it Black Pepper}, and {\it Apple}. Let us consider the two extreme cases, viz. {\it Elderberry}, for which interference is the most weakening ($\phi=-113.2431^\circ$), and {\it Mushroom}, for which it is the most strengthening ($\phi=18.6744$). For {\it Elderberry}, we have $\mu(A)=0.1138$ and $\mu(B)=0.0170$, which means that test subjects have classified {\it Elderberry} very strongly as {\it Fruits} ({\it Apple} is the most strongly classified {\it Fruits}, but {\it Elderberry} is next and close to it), and quite weakly as {\it Vegetables}. For {\it Mushroom}, we have $\mu(A)=0.0140$ and $\mu(B)=0.0545$, which means that test subjects have weakly classified {\it Mushroom} as {\it Fruits} and moderately as {\it Vegetables}. Let us suppose that $1/2(\mu(A)+\mu(B))$ is the value estimated by test subjects for `{\it Fruits or Vegetables}'. In that case, the estimates for {\it Fruits} and {\it Vegetables} apart would be carried over in a determined way to the estimate for `{\it Fruits or Vegetables}', just by applying this formula. This is indeed what would be the case if the decision process taking place in the human mind worked as if a classical particle passing through the {\it Fruits} hole or through the {\it Vegetables} hole hit the mind and left a spot at the location of one of the exemplars. More concretely, suppose that we ask subjects first to choose which of the questions they want to answer, {\it Question A} or {\it Question B}, and then, after they have made their choice, we ask them to answer this chosen question. This new experiment, which we could also indicate as {\it Question A} or {\it Question B}, would have $1/2(\mu(A)+\mu(B))$ as outcomes for the weight with respect to the different exemplars. In such a situation, it is indeed the mind of each of the subjects that chooses randomly between the {\it Fruits} hole and the {\it Vegetables} hole, subsequently following the chosen hole. There is no influence of one hole on the other, so that no interference is possible. However, in reality the situation is more complicated. When a test subject makes an estimate with respect to `{\it Fruits or Vegetables}', a new concept emerges, namely the concept `{\it Fruits or Vegetables}'. For example, in answering the question whether the exemplar {\it Mushroom} is a good example of `{\it Fruits or Vegetables}', the subject will consider two aspects or contributions. The first is related to the estimation of whether {\it Mushroom} is a good example of {\it Fruits} and to the estimation of whether {\it Mushroom} is a good example of {\it Vegetables}, i.e. to estimates of each of the concepts separately. It is covered by the formula $1/2(\mu(A)+\mu(B))$. The second contribution concerns the test subject's estimate of whether or not {\it Mushroom} belongs to the category of exemplars that cannot readily be classified as {\it Fruits} or {\it Vegetables}. This is the class characterized by the newly emerged concept `{\it Fruits or Vegetables}'. And as we know, {\it Mushroom} is a typical case of an exemplar that is not easy to classify as `{\it Fruits or Vegetables}'. That is why {\it Mushroom}, although only slightly covered by the formula $1/2(\mu(A)+\mu(B))$, has an overall high score as `{\it Fruits or Vegetables}'. The effect of interference allows adding the extra value to $1/2(\mu(A)+\mu(B))$ resulting from the fact that {\it Mushroom} scores well as an exemplar that is not readily classified as `{\it Fruits or Vegetables}'. This explains why {\it Mushroom} receives a strengthening interference effect, which adds to the probability of it being chosen as a good example of `{\it Fruits or Vegetables}'. {\it Elderberry} shows the contrary. Formula $1/2(\mu(A)+\mu(B))$ produces a score that is too high compared to the experimentally tested value of the probability of its being chosen as a good example of `{\it Fruits or Vegetables}'. The interference effect corrects this, subtracting a value from $1/2(\mu(A)+\mu(B))$. This corresponds to the test subjects considering {\it Elderberry} `not at all' to belong to a category of exemplars hard to classify as {\it Fruits} or {\it Vegetables}, but rather the contrary. As a consequence, with respect to the newly emerged concept `{\it Fruits or Vegetables}', the exemplar {\it Elderberry} scores very low, and hence the $1/2(\mu(A)+\mu(B))$ needs to be corrected by subtracting the second contribution, the quantum interference term. A similar explanation of the interference of {\it Fruits} and {\it Vegetables} can be put forward for all the other exemplars. The following is a general presentation of this. `For two concepts $A$ and $B$, with probabilities $\mu(A)$ and $\mu(A)$ for an exemplar to be chosen as a good example of `$A$ or $B$', the interference effect allows taking into account the specific probability contribution for this exemplar to be chosen as a good exemplar of the newly emerged concept `$A\ {\rm or}\ B$', adding or subtracting to the value $1/2(\mu(A)+\mu(B))$, which is the average of $\mu(A)$ and $\mu(B)$.'

To conclude we observe that `{\it Fruits or Vegetables}' is not the only case where quantum interference explains deviations from classically expected behavior. Various examples have been found, for disjunctions, as well as for conjunctions, of concepts \cite{aerts2009}.

\section{A two-layered structure in human thought\label{layers}}
The detection of quantum structures in cognition has suggested us to put forward the hypothesis that two specifically structured and superposed layers can
be identified in human thought as a process \cite{aerts2009,aertsdhooghe2009}, as follows.

(i) A {\it classical logical layer}. The thought process in this layer is given form by an underlying classical logical conceptual process. The manifest process itself may be, and generally will be, indeterministic, but the indeterminism is due to a lack of knowledge about the underlying deterministic classical process. For this reason the process within the classical logical layer can be modeled by using a classical Kolmogorovian probability description.

(ii) A {\it quantum conceptual layer}. The thought process in this layer is given form under influence of the totality of the surrounding conceptual landscape, where the different concepts figure as individual entities, also when they are combinations of other concepts, at variance with the classical logical layer where combinations of concepts figure as classical combinations of entities and not as individual entities. In this sense one can speak of a {\it conceptual emergence} taking place in this quantum conceptual layer, certainly so for combinations of concepts. Quantum conceptual thought has been identified in different domains of knowledge and science related to different, often as paradoxically conceived, problems in these domains. The sorts of measurable quantities being able to experimentally identify quantum conceptual thought have been different in these different domains, depending on which aspect of the conceptual landscape was most obvious or most important for the identification of the deviation of classically expected values of these quantities. For example, in a domain of cognitive science where representations of concepts are studied, and hence where concepts and combinations of concepts, and relations of items, exemplars, instances or features with concepts are considered, measurable quantities such as `typicality', `membership', `similarity' and `applicability' have been studied and used to experimentally put into evidence the deviation of what classically would be expected for the values of these quantities. In decision theory measurable quantities such as `representativeness', `qualitative likelihood', `similarity' and `resemblance' have played this role. The quantum conceptual thought process is indeterministic in essence, i.e. there is not necessarily an underlying deterministic process independent of the context. Hence, if analyzed deeper with the aim of finding more deterministic sub-processes, unavoidably effects of context will come into play. Since all concepts of the interconnected web that forms the landscape of concepts and combinations of them attribute as individual entities to the influences reigning in this landscape, and more so since this happens dynamically in an environment where they are all quantum entangled structurally speaking, the nature of quantum conceptual thought contains aspects that we strongly identify as holistic and synthetic. However, the quantum conceptual thought process is not unorganized or irrational. Quantum conceptual thought is as firmly structured as classical logical thought though in a different way. We believe that the reason why science has hardly uncovered the structure of quantum conceptual thought is because it has been believed to be intuitive, associative, irrational, etc., meaning `rather unstructured'. As a consequence of its basic features, an idealized version of this quantum conceptual thought process can be modeled as a quantum mechanical process. 

The assumed existence of a quantum conceptual layer in mind fits in with some impressive achievements that have been recently obtained in neuroscience \cite{brain2012}, as we will see in the next section.

\section{Quantum cognition and the structure of the brain\label{brain}}
A traditional view of the relation between brain and mind is based on the {\it neuroscience paradigm} \cite{paradigm}, according to which the architecture of the brain is determined by connections between neurons, their inibitory/excitatory character, and the strength of their connections. Following this view, roughly speaking, the brain can be seen as a {\it parallel distributed computer} containing many billions of neurons, that is, elementary processors interconnected into a complex neural network. In this architecture, the mind and the brain constitute one single unit, which is characterized by a complementary dualism. The mind is in this approach understood as a program carried out by the brain, the program being specified by the neural network architecture.  Distributed representations of cognitive structures are studied in such an approach (see, e.g., {\it holographic reduced representations} \cite{gabor1968}--\cite{plate2003}).

Although the holographic approach is inspired by waves and interference, it is not able to model the complex type of interference that quantum entities undergo. It can be seen by considering the values of the interference angles of the interference pattern we obtain (see equation (\ref{interferenceangles})), that the modeling for the concept {\it Fruit or Vegetables} is intrinsically quantum mechanical, not able to be reduced to interference of classical waves. This means that, although along the same lines as the holographic memory view \cite{gabor1968}, our approach can introduce a way to consider and study the brain as a quantum mechanical interference producing entity. Concretely we produce a projection of a multi--dimensional complex Hilbert space -- 25 dimensional for the {\it Fruits or Vegetables} case -- in three--dimenesional real space, which is the environment where the bio-mass of the brain is located. 

In this respect it is worthy to mention a recent finding \cite{brain2012}, where relationships of adjacency and crossing between cerebral fiber pathways in primates and humans were analyzed by using diffusion magnetic resonance imaging. The cerebral fiber pathways have been found to form a rectilinear three-dimensional grid continuous with the three principal axes of development. Cortico-cortical pathways formed parallel sheets of interwoven paths in the longitudinal and medio-lateral axes, in which major pathways were local condensations. Cross-species homology was strong and showed emergence of complex gyral connectivity by continuous elaboration of this grid structure. This architecture naturally supports functional spatio-temporal coherence, developmental path-finding, and incremental rewiring with correlated adaptation of structure and function in cerebral plasticity and evolution \cite{brain2012}. The three--dimensional layered structure schematized above puts at stake the `neural network' modeling of the brain, together with some aspects of the neuroscience paradigm, and the brain/mind relation. Such a very mathematically structured grid form would be much closer to what one expect as an ideal medium for interference than this is the case for the structure of a traditional network.
 
At first sight it might seem that the layered structures that have been detected \cite{brain2012} are too simple to give rise to complex cognition, even if interference is allowed to play a prominent role, but that is misleading. Indeed, one should not look upon the brain as `a container of complex cognition', but rather as `the canvas for the potentiality of emergence of such complex cognition'. That makes a whole difference. Indeed, we know how the rather simple mathematical structure of superposition in a linear vector space and tensor product of linear vector spaces give rise to both emergence and entanglement in quantum mechanics. Also there this mathematical structure plays the role of canvas, where the emergent and entangled states can find a seat to be realized. This is exactly what the role of the recently detected grid could be, due to its rather simple mathematical structure, at least compared to the structure of a network, it could make available in a mathematically systematic way the canvas where emergent states of new concepts can find their seat. This is then a mechanism fundamentally different  from what one expects in networks, where `new connections are only made when they are needed'. Structures that have generative power can shape `empty space' for potentiality, and `creation of new', hence emergence can take place in a much more powerful way. Of course, there will be a bias coming from the generating structures, which is a drawback compared to the network way. This bias could exactly be an explanation for the functioning of the human brain leading to automated aspects of conceptual reasoning such as `the disjunction and conjunction effects'. The above analysis is highly relevant for representations of genuine cognitive models in technology, for example as attempted in artificial intelligence and robotics \cite{penrose1990}--\cite{dongchenzhangchen2006}.

\end{document}